\documentclass{article}

\PassOptionsToPackage{numbers, compress}{natbib}

\usepackage[main,final]{neurips_2026}

\usepackage[utf8]{inputenc} 
\usepackage[T1]{fontenc}    
\usepackage{hyperref}       
\usepackage{url}            
\usepackage{booktabs}       
\usepackage{amsfonts}       
\usepackage{nicefrac}       
\usepackage{microtype}      
\usepackage{xcolor}         

\usepackage{graphicx}
\usepackage{amsmath}
\usepackage{amsthm}
\usepackage{booktabs}
\usepackage{algorithm}
\usepackage{algorithmic}
\usepackage{multirow}
\usepackage[caption=false]{subfig}
\usepackage{amssymb}
\usepackage{booktabs}
\usepackage{color}
\usepackage{colortbl}
\usepackage{makecell}
\usepackage{array}
\usepackage{pifont}       
\usepackage{bbding}  
\usepackage{wrapfig}

\definecolor{mygray}{gray}{.9}
\usepackage{makecell} 
\usepackage{etoolbox}
\definecolor{lightgray}{gray}{0.92}

\title{Masks Can Talk: Extracting Structured Text Information from Single-Modal Images for Remote Sensing Change Detection}

\author{%
  Kai Zheng$^{\dagger,1}$, Hang-Cheng Dong$^{\dagger,2,3}$, Jiatong Pan$^{1}$, Zhenkai Wu$^{1}$ Fupeng Wei$^{4,5}$, Wei Zhang$^{1}$\thanks{Corresponding author.}\thanks{Equal contribution.}  \\
  $^{1}$Zhejiang University\\
  $^{2}$Harbin Institute of Technology\\
  $^{3}$Harbin Institute of Technology Suzhou Research Institute\\
  $^{4}$North China University of Water Resources and Electric Power\\
  $^{5}$University of Auckland\\
  \texttt{zhengkai1990@zju.edu.cn, hunsen\_d@hit.edu.cn,cstzhangwei@zju.edu.cn} \\
}
  
\begin{document}


\maketitle

\begin{abstract}
Remote sensing change detection is pivotal for urban monitoring, disaster assessment, and environmental resource management. Yet, unimodal deep learning methods frequently confuse genuine semantic changes with visually similar but irrelevant variations. Recent multimodal approaches incorporate text as auxiliary supervision, but their descriptions are either semantically coarse and unstructured or model-generated and thus noisy. Critically, all of them overlook a simple fact: fine-grained change semantics are already implicitly encoded in the ground-truth mask labels that come standard with every change detection dataset. These masks know where the change happened, what the land-cover types were before and after, how the transition occurred, and how many objects were involved. In this paper, we propose S2M, a framework that obtains structured textual features directly from change labels at zero additional annotation cost. Specifically, each change region is automatically transcribed into a semantic quadruple (where, what, how, how many) and converted into several fixed-template text descriptions, providing precise, dense, and noise-free multimodal supervision. We adopts a two-stage training strategy to fine-tune on remote sensing imagery firstly for robust domain-specific representation, after which a multimodal decoder with a bi-directional contrastive loss is introduced to achieve deep alignment between visual features and structured textual embeddings. To validate our method, we construct Gaza-Change-v2, a new multi-class change detection (MCD) dataset about the Gaza Strip. On this MCD dataset, S2M achieves a Sek of 17.80\% and an F$_{\text{scd}}$ of 66.14\%, notably surpassing even multimodal methods that leverage large language models. Our work demonstrates that masks can indeed talk. They tell us exactly what, where, how, and how many changes have occurred.
\end{abstract}


\section{Introduction}
\vspace{-3mm}

Remote sensing change detection (RSCD) technology facilitates decision-making support for critical applications such as urban monitoring, disaster damage assessment, and environmental protection~\cite{isprs,holail2025edge}. The core challenge of RSCD lies in distinguishing semantically meaningful changes from visually confusing pseudo-changes. Deep learning methods, whether based on convolutional neural networks (CNN)~\cite{FCsiam2018} or vision Transformers (VIT)~\cite{changeformer}, have significantly advanced the field by learning increasingly powerful visual representations. However, when semantically distinct changes such as "farmland turning into bare soil" and "building destruction" produce highly similar patterns in visual feature space, pure vision models lack the semantic prior information necessary to reliably separate them. This inter-class appearance similarity, prevalent in aerial imagery, sets a hard ceiling on what unimodal perception can achieve.


The recent rise of vision-language models (VLMs), particularly CLIP~\cite{radford2021learning}, has inspired a multimodal turn in the field of change detection. By introducing text as a complementary modality, several pioneering works have demonstrated that linguistic knowledge can help disambiguate visually confusing events and suppress pseudo-changes. ChangeCLIP~\cite{dong2024changeclip} leverages CLIP's zero-shot classification capability to generate category-level text prompts. MMChange~\cite{MMchange} and UniChange~\cite{zhang2025unichange} employ trainable VLMs and multimodal large language models (MLLMs)~\cite{liu2023visual,zhu2023minigpt,dai2023instructblip}, respectively, to generate richer natural language descriptions of entire scenes. TextMCD~\cite{wu2026textmcd} explores the use of text embeddings as auxiliary supervisory signals to regularize mask-level pr edictions. However, existing multimodal methods rely on the acquisition of textual description labels, which face the following challenges: (1) manually adding text label descriptions is time-consuming, labor-intensive, and costly; (2) labels obtained from VLMs are noisy, inaccurate, and overly dependent on the performance of the VLM itself; (3) high-performing MLLMs possess enormous parameter counts, introducing substantial training overhead. Especially in practical application scenarios, the constraints of computational resources and human effort greatly limit the applicability of multimodal remote sensing change detection.

In this work, we observe that the ground-truth mask labels already present in every change detection dataset contain rich, fine-grained semantic information about change events. These masks implicitly encode the situations before and after a change, reveal where the change occurred, and even indicate how many distinct changed objects are involved. Yet, astonishingly, this wealth of structured semantic knowledge has never been explicitly verbalized and leveraged as multimodal supervision.




Based on the above insights, we propose a method that upgrades unimodal data to multimodal data at zero cost, named S2M. Specifically, we argue that the textual information inherently contained in change mask labels can be represented as a textual composition $\mathcal{T}$ consisting of a quadruple. Then the elements of a classical textual composition $\mathcal{T}(L,C,T,Q)$ are: 
\begin{itemize}
    \item Where/$[$location$]$: the spatial location of the changed region;
    \item  What/$[$category$]$: the land-cover categories before and after the change;

    \item How/$[$type$]$: the type of transition (e.g., "built", "destroy");

    \item How many/$[$quantity$]$: the number of changed object instances.
\end{itemize}

\begin{figure}[t]
	\centering
	\includegraphics[width=1\textwidth]{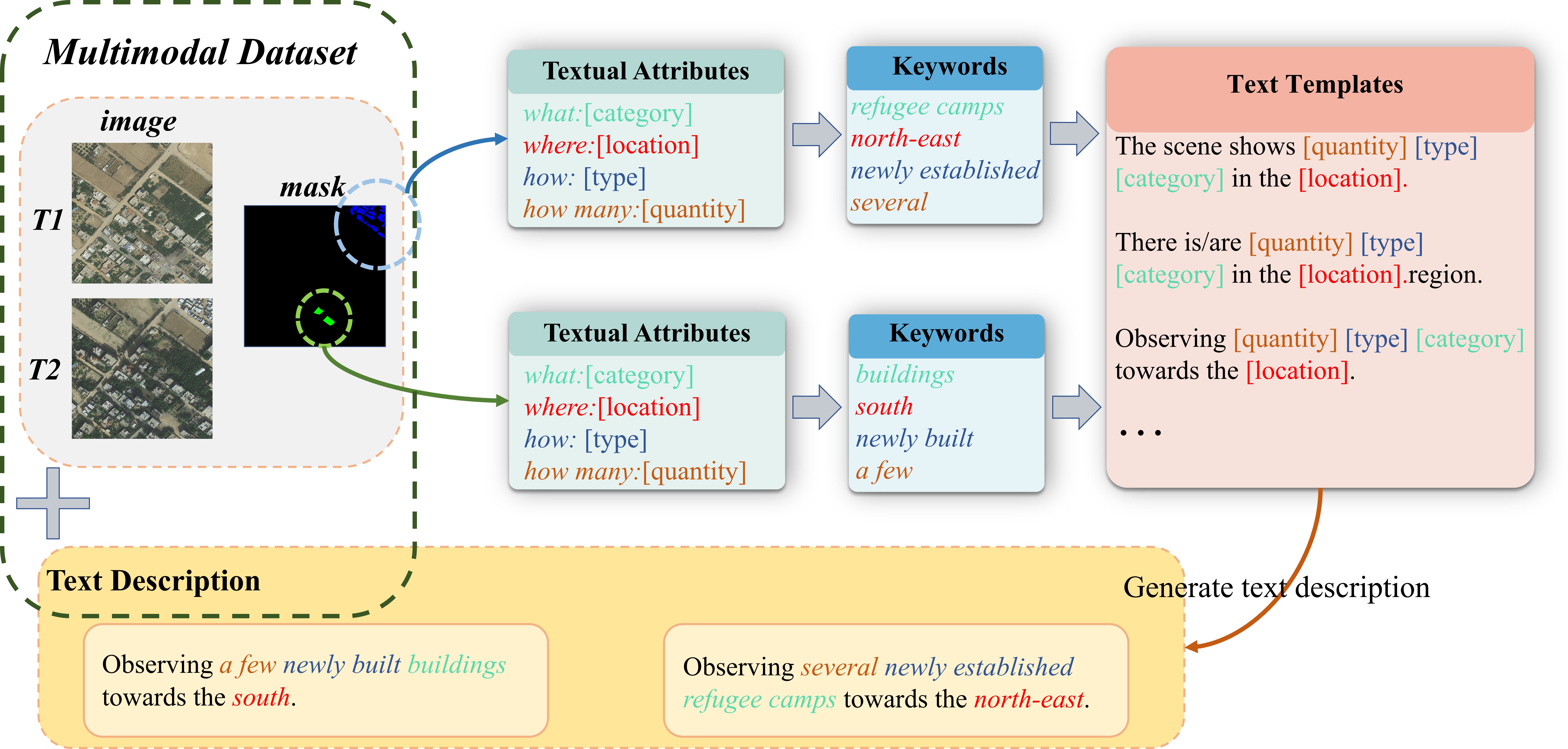}
	\caption{Flowchart of the proposed S2M method. Textual features are generated from conventional change detection data, incorporating four elements: what, where, how, and how many, which correspond respectively to the mask's category, the location of changed regions, the type of change, and the quantity of change.}
	\label{fig:textgenerate}
 \vspace{-4mm}
\end{figure}

S2M automatically extracts these textual composition $\mathcal{T}(l,c,t,q)$ and instantiates them through several fixed templates. The text extraction pipeline is illustrated in Figure \ref{fig:textgenerate}. Such descriptions are structured, precise, dense, noise-free, and perfectly aligned with change regions. Critically, this entire process incurs zero additional annotation cost. We argue that S2M textualizes the image information already known by the masks, and this cross-modal form of information provides cross-modal prior guidance for the model to distinguish visually similar yet semantically distinct features, thereby helping the model learn fine-grained inter-class feature differences.

To effectively exploit these structured textual features, we carefully design a two-stage training pipeline. In the first stage, the visual backbone is fine-tuned on remote sensing images to learn robust domain-specific visual representations. In the second stage, we introduce a text-guided multimodal alignment decoder that fuses the structured text embeddings with multi-scale visual features. During the second stage, a bidirectional vision-text contrastive loss based on~\cite{oord2018representation} is introduced to align visual representations and their corresponding textual descriptions in a shared embedding space. This bidirectional alignment ensures that the model learns semantic priors, thereby enlarging the inter-class distance between representations of visually similar yet semantically distinct features, and addressing the inter-class similarity challenge that unimodal methods cannot overcome.

Meanwhile, we construct a dataset tailored to real-world complex application scenarios, named Gaza-Change-v2, to validate the effectiveness of the S2M method. This dataset is a multi-class remote sensing change detection dataset focusing on infrastructure changes in the Gaza Strip between 2023 and 2024. It encompasses diverse change types during the conflict, including building destruction and reconstruction, temporary shelter construction, and farmland damage. In addition, we conduct comprehensive experiments on standard binary change detection benchmarks (LEVIR-CD and WHU-CD).

In summary, our contributions are threefold:

\begin{itemize}

\item     We analyze the practical limitations of existing multimodal remote sensing change detection methods and propose S2M, a textual generation scheme that upgrades unimodal data to multimodal data without introducing additional manual annotation or reliance on large language models. This "free information lunch" substantially enhances model performance in real-world remote sensing detection scenarios.

\item      We design a bidirectional vision-text contrastive loss to align textual information with visual information, enhancing the model's ability to resolve inter-class similarity issues among visual features.

\item      We construct a new multi-class remote sensing change detection dataset, Gaza-Change-v2. Furthermore, we conduct experiments on widely adopted benchmarks LEVIR-CD and WHU-CD, where the proposed scheme consistently achieves state-of-the-art results across all datasets. 

\end{itemize}

\section{Related Work}
\vspace{-3mm}

\textbf{Visual-only remote sensing change detection.} Existing visual change detection methods can be divided into convolution‑based and transformer‑based categories. Both take bi‑temporal images as input and employ siamese network architectures~\cite{FCsiam2018,zhan2017change} to extract bi‑temporal features. By enhancing contextual feature modeling, these methods can improve feature fusion and extraction, achieving reliable detection under complex environmental backgrounds~\cite{wu2025cdxlstm, shen2022semantic,liu2020building, attention}. Transformer‑based methods model global spatiotemporal dependencies to achieve superior performance, thereby surpassing convolution‑based approaches. BIT~\cite{BIT} fuses Transformer and CNN structures, fully exploiting the complementary nature of convolution and patch‑based feature extraction. In contrast, ChangeFormer~\cite{changeformer} adopts a pure Transformer framework, while Swin‑SUNet~\cite{zhang2022swinsunet} introduces the more advanced Swin Transformer~\cite{liu2021swin} into the remote sensing change detection task. Despite these advances, purely visual methods operate within a single modality, lacking explicit semantic guidance to distinguish semantically distinct yet visually similar change types. Particularly in sudden scenarios such as natural disasters or building damage, the high inter‑class similarity within the single image modality severely limits the practical performance of deep learning methods.

\textbf{Text‑guided multimodal change detection.} To overcome the semantic bottleneck of visual-only methods, recent studies introduce natural language as an auxiliary modality. ChangeCLIP~\cite{dong2024changeclip} reconstructs CLIP~\cite{radford2021learning} to symmetrically process bi-temporal images and text, adding a difference compensation module and a vision-language decoder. TextMCD~\cite{wu2026textmcd} also uses CLIP to encode multi-class visual mask features, providing richer semantic embeddings for visual features. Semantic-CD~\cite{11243524} adopts a similar design with an open-vocabulary prompter. MMChange~\cite{zhou2025multimodal} employs a VLM to generate corresponding text descriptions and designs an image-text feature fusion module. UniChange~\cite{zhang2025unichange} uses a multimodal large language model to embed text features. However, existing methods heavily rely on image-to-text transformation, which imposes high demands on training and deployment. In contrast, our work directly leverages the textual features naturally inherent in images and labels, generating descriptive text according to predefined rules. This text modality generation approach significantly reduces training overhead and offers a new solution to the inter-class similarity problem in visual images.

\section{Methodology}
\vspace{-3mm}

\subsection{S2M: Text Descriptor Generation}
\label{s2m}
The cornerstone of S2M is the automatic conversion of silent change masks into structured textual descriptions.
We argue that every change event delineated by a ground‑truth mask can be exhaustively described by a semantic quadruple of four fundamental elements:
\textit{where} (spatial location),
\textit{what} (land‑cover category),
\textit{how} (change action), and
\textit{how many} (object quantity).
The following extraction rules are specifically tailored to the Gaza‑Change‑v2 dataset. The same principle generalizes naturally to any change detection dataset with categorical annotations.

\begin{figure}[ht]
	\centering
	\includegraphics[width=1\textwidth]{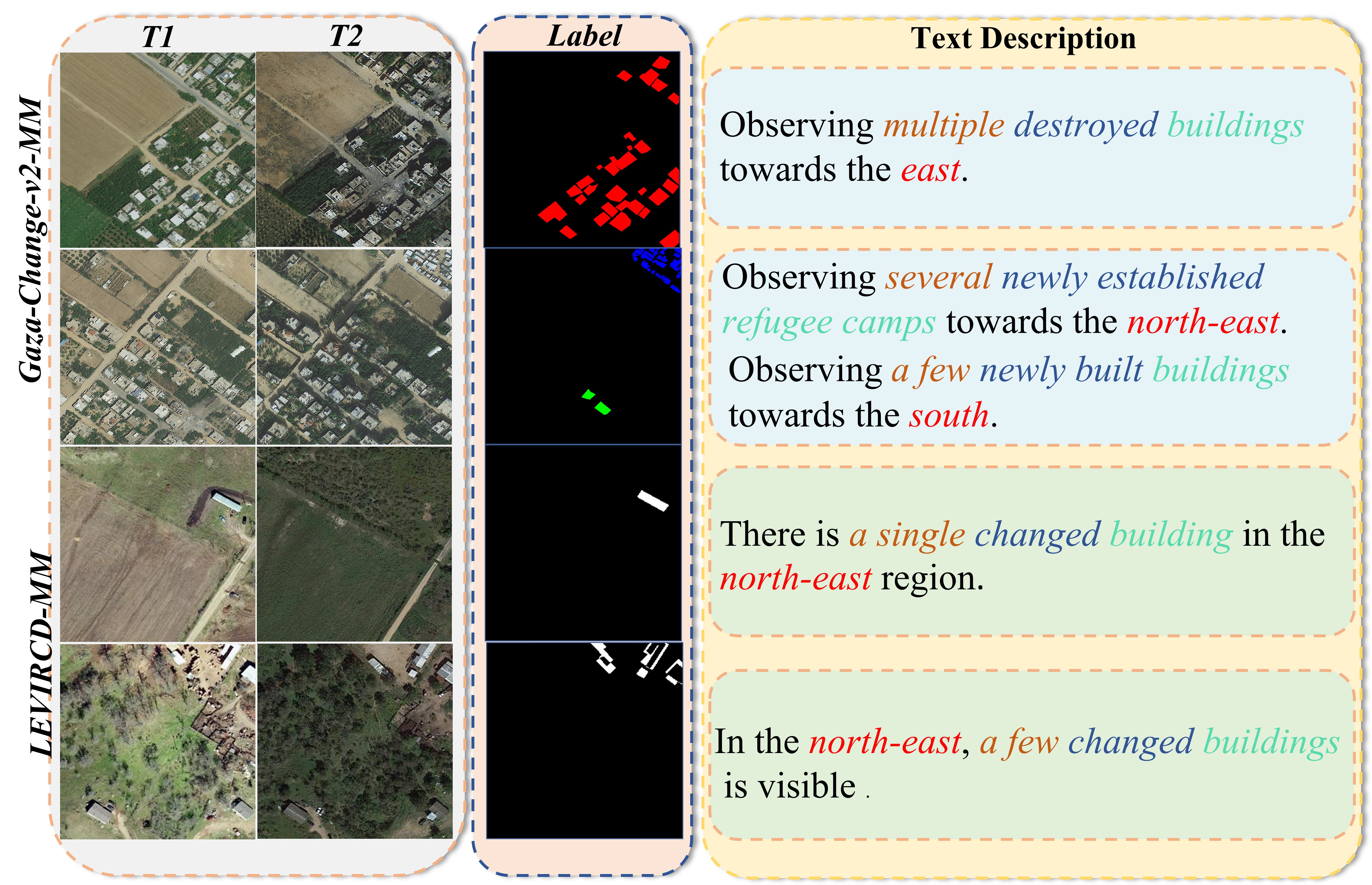}
    \vspace{-5mm}
	\caption{Visualization of the constructed multimodal change detection dataset. The top two rows correspond to the Gaza-Change dataset, which contains multiple change categories, while the bottom two rows correspond to the LEVIR-CD dataset, which has only two categories. The new multimodal dataset, named by appending the suffix “-MM” to the original dataset name.}
	\label{fig:dataintro}
 \vspace{-4mm}
\end{figure}

\subsubsection{Definition of the semantic quadruple}
The four elements are defined as follows:
\begin{itemize}
    \item \textbf{Direction}~($D$): the spatial location of the changed region’s centroid within the image, discretized into nine orientation categories: \textit{north}, \textit{south}, \textit{east}, \textit{west}, \textit{center}, \textit{northeast}, \textit{northwest}, \textit{southeast}, \textit{southwest}.
    \item \textbf{Quantity}~($Q$): the approximate number of changed objects, categorized into four granularity levels based on pixel area: \textit{a single}, \textit{a few}, \textit{several}, \textit{multiple}.
    \item \textbf{Category}~($C$): the land‑cover type of the changed object, e.g., \textit{buildings}, \textit{building}, \textit{refugee camp}, \textit{agricultural land}, \textit{greenhouse destroyed}, \textit{greenhouse newly built}.
    \item \textbf{Type}~($T$): the change type that occurred, e.g., \textit{destroyed}, \textit{newly built}, \textit{newly established}.
\end{itemize}

\subsubsection{Direction computation}
Given a binary change mask of size $H \times W$, we first compute its centroid by averaging the spatial coordinates of all foreground pixels:
\begin{equation}
    c_x = \frac{1}{N}\sum_{i=1}^{N} x_i, \quad c_y = \frac{1}{N}\sum_{i=1}^{N} y_i,
\end{equation}
where $N$ is the number of changed pixels.
The image is conceptually partitioned into a $3\times 3$ spatial grid.
The vertical direction $d_v$ is determined as:
\begin{equation}
    d_v = \begin{cases}
        \text{north}, & c_y < H/3 \\
        \text{center}, & H/3 \leq c_y < 2H/3 \\
        \text{south}, & c_y \geq 2H/3
    \end{cases}
\end{equation}
The horizontal direction $d_h$ is analogously defined with thresholds at $W/3$ and $2W/3$, yielding \textit{west}, \textit{center}, or \textit{east}.
The final direction $D$ is obtained by combining $d_v$ and $d_h$.

\subsubsection{Quantity determination}
The quantity descriptor $Q$ is determined by the total pixel area of the change mask using empirically calibrated thresholds:
\begin{equation}
    Q = \begin{cases}
        \text{a single}, & N < 800 \\
        \text{a few},    & 800 \leq N < 4000 \\
        \text{several},  & 4000 \leq N < 8000 \\
        \text{multiple}, & N \geq 8000
    \end{cases}
\end{equation}

\subsubsection{Template-based text generation}
\setlength{\tabcolsep}{3pt}
\begin{wraptable}{r}{0.5\linewidth}
\scriptsize
	\centering
 \vspace{-5pt}
		\caption{
		Text templates used in our method.
		}
  \vspace{3pt}
		\label{tablb:texttemplate}
            \begin{tabular}{c||c}
		\Xhline{1.2pt}
            \rowcolor{mygray}
		       No. &Template \\	
                \hline 
                \hline
                 \textcircled{1}& The scene shows $[$quantity$]$ $[$type$]$ $[$category$]$ in the $[$location$]$.\\
                 \textcircled{2} & Observing $[$quantity$]$ $[$type$]$ $[$category$]$ towards the $[$location$]$.\\
                 \textcircled{3} & $[$quantity$]$ $[$type$]$ $[$category$]$ located in the $[$location$]$.\\
                 \textcircled{4} & In the $[$location$]$, $[$quantity$]$ $[$type$]$ $[$category$]$ is/are visible.\\
                 \textcircled{5}& There is/are $[$quantity$]$ $[$type$]$ $[$category$]$ in the $[$location$]$ region  \\
   			\hline
		\end{tabular}
  \vspace{-10pt}
\end{wraptable}

Once the four elements are instantiated, they are assembled into a natural language description using a fixed template.
To increase linguistic diversity and prevent the model from overfitting to specific phrasings, we design five distinct templates and randomly select one for each training sample, as represented in Table \ref{tablb:texttemplate}.
For the special case where $D = \text{center}$, a neutral variant is used:
\texttt{The scene shows $\langle Q\rangle$ $\langle T\rangle$ $\langle C\rangle$ in the center.}

\subsection{Model Architecture and Training Strategy}
\label{sec:architecture}
\begin{figure}[t]
	\centering
	\includegraphics[width=1\textwidth]{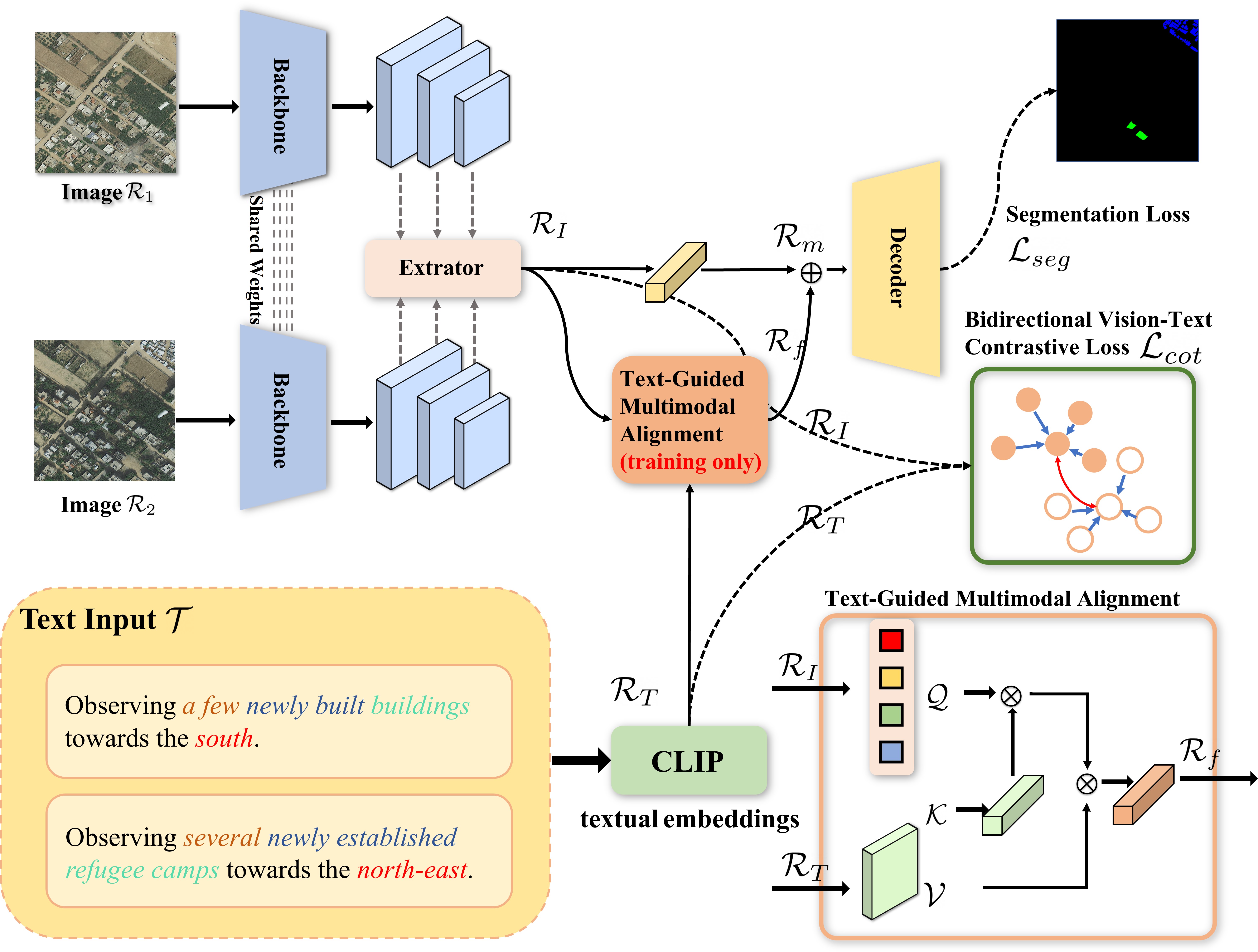}
    \vspace{-5mm}
	\caption{The proposed model training pipeline. It is worth mentioning that the introduced text features are used only during training. At the inference stage, neither the text nor the corresponding fusion module is required.}
	\label{fig:method}
 \vspace{-4mm}
\end{figure}

We employs a two‑stage training paradigm that progressively incorporates structured textual supervision into a high‑capacity visual change detector.

\subsubsection{Stage 1: Visual Pre‑training with MC‑DiSNet}

In the first stage, we build upon the MC‑DiSNet architecture~\cite{isprs}, which leverages DINOv3~\cite{simeoni2025dinov3} as its visual backbone. MC‑DiSNet has been shown to capture fine‑grained spatio‑temporal change patterns. We streamline the encoder of MC‑DiSNet to better accommodate the subsequent multimodal fusion. We introduce a feature bridge $\mathbf{W}_b$ between the encoder and the decoder. This bridge serves as a gated connection that adaptively regulates how textual semantics modulate the multi‑scale visual representations before they enter the decoder, ensuring that the cross‑modal signal complements rather than overrides the fine‑grained spatial features learned during Stage~1.

\subsubsection{Stage 2: Text‑Guided Multimodal Alignment}
Starting from the weights initialized in the first stage, we perform re-joint training by introducing text inputs \(\mathcal{T}\). As shown in Figure \ref{fig:method}, the text descriptions are encoded by a frozen pre-trained CLIP text encoder to obtain text embeddings \(\mathcal{R}_T\). To bridge the cross-modal gap, we design a \textbf{Text-Guided Multimodal Alignment} module, which performs cross-attention with visual features as queries ($\mathcal{Q}$) and text features as keys ($\mathcal{K}$)) and values ($\mathcal{V}$). Then, the multimodal feature \(\mathcal{R}_m\) is calculated by

\begin{equation}
   \mathcal{R}_m = Attention(\mathcal{Q},\mathcal{K},\mathcal{V}) + \mathbf{W}_b\mathcal{R}_I,
\end{equation}
where $Attention$ is a standard attention mechanism. During this stage, the entire network, including the visual backbone and decoder, is jointly optimized end-to-end.

\subsection{Loss Functions}
\label{sec:loss}
The entire framework is jointly optimized with two complementary losses: (1) a standard \textbf{segmentation loss} \(\mathcal{L}_{\text{seg}}\) applied to the final decoder output to supervise the change mask prediction; and (2) a \textbf{bidirectional vision-text contrastive loss} \(\mathcal{L}_{\text{cot}}\) defined in the feature space. The contrastive loss explicitly forces semantic consistency between the visually extracted features \(\mathcal{R}_I\) and the corresponding textual embeddings \(\mathcal{R}_T\), thereby encouraging the visual encoder to mine fine-grained cross-modal features during training. 

\textbf{Segmentation loss.} We design a composite loss landscape that jointly addresses class imbalance, boundary refinement, and multimodal alignment. For the segmentation task, we adopt a hybrid segmentation loss combining \textbf{Focal Loss}~\cite{focalloss}, \textbf{Dice Loss}~\cite{diceloss}, and \textbf{Lovász Loss}~\cite{lovaszloss}.
The segmentation loss is defined as:

\begin{equation}
    \mathcal{L}_{\text{seg}} = \alpha\mathcal{L}_{\text{focal}} + \beta\mathcal{L}_{\text{dice}} + \gamma\mathcal{L}_{\text{lovász}}.
\end{equation}

\textbf{Bidirectional vision‑text contrastive loss.}
To align the visual and textual modalities, we employ a bidirectional infoNCE loss based on \cite{he2020momentum}.
Let \(\mathcal{R}^i_I\) be the global visual feature of sample~$i$, and \(\mathcal{R}^i_T\) the corresponding text embedding.
A similarity matrix $\mathbf{S}_{ij} = \nicefrac{\mathcal{R}_I^i \cdot \mathcal{R}^j_T}{\tau}$ is computed with temperature $\tau$.
The vision‑to‑text ($\text{V}\!\rightarrow\!\text{T}$) and text‑to‑vision ($\text{T}\!\rightarrow\!\text{V}$) losses are:

\begin{equation}
    \mathcal{L}_{\text{V}\!\rightarrow\!\text{T}} = -\frac{1}{B}\sum_{i=1}^{B} \log \frac{\exp(\mathbf{S}_{ii})}{\sum_j \exp(\mathbf{S}_{ij})},
    \quad
    \mathcal{L}_{\text{T}\!\rightarrow\!\text{V}} = -\frac{1}{B}\sum_{i=1}^{B} \log \frac{\exp(\mathbf{S}_{ii})}{\sum_j \exp(\mathbf{S}_{ji})}.
\end{equation}

The bidirectional contrastive loss is their average:

\begin{equation}
    \mathcal{L}_{\text{cot}} = \frac{1}{2}\bigl(\mathcal{L}_{\text{V}\!\rightarrow\!\text{T}} + \mathcal{L}_{\text{T}\!\rightarrow\!\text{V}}\bigr).
\end{equation}

\textbf{Overall training objective.}
The total loss for Stage~2 integrates the pixel‑level segmentation supervision with the multimodal alignment signal:

\begin{equation}
    \mathcal{L}_{\text{total}} = \mathcal{L}_{\text{seg}} + \lambda \cdot \mathcal{L}_{\text{cot}},
\end{equation}

where $\lambda$ is a balancing hyperparameter. This combined objective encourages the model to produce accurate change maps while learning semantic, modality‑invariant representations that are discriminative across visually similar change classes.

\section{Experiments}
\vspace{-3mm}
We evaluate the proposed S2M on three remote sensing change detection datasets: \textbf{Gaza‑Change‑v2}, \textbf{LEVIR‑CD}~\cite{levir}, and \textbf{WHU‑CD}~\cite{whu}. Detailed descriptions and statistics of these datasets are provided in Appendix~\ref{sec:A} (Table~\ref{table:dataset3info}). All experiments are implemented in PyTorch and conducted on a single NVIDIA RTX 3090 Ti GPU. We train all models for 150 epochs, with validation performed at the end of each epoch. The AdamW optimizer is used with an initial learning rate of $1\times 10^{-4}$ and a weight‑decay coefficient of $0.05$. Hyperparameters $\alpha$, $\beta$, and $\gamma$ are set to $0.4$, $0.3$, and $0.3$, respectively, and $\lambda$ is set to $0.5$. The temperature $\tau$ in the loss function is set as 0.7~\cite{radford2021learning}.  Evaluation metrics follow the reference paper~\cite{ding2022bi}.
Further implementation details and hyperparameter settings can be found in the Appendix.

\subsection{Main Results}

\textbf{Results on Gaza‑Change‑v2‑MM.} Table~\ref{table:maintab} reports the quantitative comparison on the Gaza‑Change‑v2‑MM dataset. Among unimodal methods ($\mathcal{S}$), MC‑DiSNet yields the best overall performance, while multimodal methods ($\mathcal{M}$) such as ChangeCLIP and TextMCD bring heavy parameter overhead with only marginal gains.
Our S2M ($\mathcal{S}\!\rightarrow\!\mathcal{M}$) consistently surpasses all competing approaches.
Specifically, S2M~(DINOv3) achieves the highest Sek of $\mathbf{17.80\%}$ and F$_{\text{scd}}$ of $\mathbf{66.14\%}$, outperforming MC‑DiSNet by $+2.03$ Sek and ChangeCLIP by $+4.68$ Sek. Notably, the LoRA‑finetuned variant S2M~(DINOv3+LoRA) attains a mIoU of $\mathbf{66.91\%}$ with only $\mathbf{7.71}$M trainable parameters.

\textbf{Results on LEVIRCD‑MM and WHUCD-MM.}On LEVIR‑CD, S2M~(DINOv3) attains the highest F$_1$ of $\mathbf{90.98\%}$ and IoU of $\mathbf{83.46\%}$, surpassing both MC‑DiSNet ($+0.26\%$ F$_1$, $+0.44\%$ IoU) and CDxLSTM ($+0.06\%$ F$_1$).
On WHU‑CD, S2M~(DINOv3+LoRA) delivers substantial gains, achieving the best F$_1$ of $\mathbf{93.49\%}$ and IoU of $\mathbf{87.78\%}$, outperforming MC‑DiSNet by $+0.70\%$ F$_1$ and $+1.23\%$ IoU. These improvements, though modest in absolute terms due to the already high baseline, are consistent across both datasets and all metrics, demonstrating that structured text provides complementary semantic cues that benefit even well‑solved BCD tasks.
\setlength{\tabcolsep}{4pt}
\begin{table*}[ht]
\scriptsize
	\begin{center}
		\caption{
        Performance comparison of RSCD on the Gaza-Change-v2-MM dataset.  All scores are reported as percentages, and the highest ones are highlighted in bold. FLOPs are computed for an input image size of $512 \times 512 \times 3$. 
		}
  \vspace{1mm}
		\label{table:maintab}
            \begin{tabular}{c||c||ccc||cccccc}
		\Xhline{1.2pt}
            \rowcolor{mygray}
		    & & & & & \multicolumn{6}{c}{Gaza-Change-v2-MM} \\
            \rowcolor{mygray}
			\multicolumn{1}{c||}{\multirow{-2}{*}{Method}}
               & \multicolumn{1}{c||}{\multirow{-2}{*}{modality}}
               & \multirow{-2}{*}{Params (M)}
               & \multirow{-2}{*}{\makecell[l]{ Trainable\\ Params (M)} }
               & \multirow{-2}{*}{Flops (G)}
               & Sek & F$_{\text{scd}}$ & mIoU & Pre. & Rec. & mF$_1$ \\			
                \hline \hline
   BIT \cite{BIT} & \multirow{13}{*}{$\mathcal{S}$} &11.89&11.89&35.38&14.36&62.48&63.76&86.90&70.37&75.91\\
   ChangeFormer~\cite{changeformer}   & &41.09&41.09&814.43&12.34&60.88&54.90&85.40&60.27&68.68\\
   CDxLSTM~\cite{wu2025cdxlstm}        & &16.18&16.18&15.45&15.45&62.84&60.30&86.02&64.74&72.56\\
   DDLNet~\cite{ddlnet}         & &13.78&13.78&29.44&13.58&62.75&58.80&86.62&62.21&71.80\\
   SNUNet~\cite{snunet}         & &12.04&12.04&219.50&12.84&61.80&56.85&84.59&61.71&69.73\\
   STNet \cite{STNet}         & &14.62&14.62&38.46&16.58&64.81&64.08&87.66&70.42&76.66\\
   USSFCNet \cite{usscfnet}      & &1.52&1.52&19.49&12.55&61.58&51.66&80.33&52.72&62.83\\
   LGPNet  \cite{LGPNet}       & &70.99&70.99&503.14&11.29&59.48&61.13&89.28&64.02&74.20\\
   Changemamba \cite{changemamba}   & &48.56&48.56&38.49&17.50&65.33&51.13&66.13&67.42&65.45\\
   RSmamba \cite{RSmamba}   & &51.95&51.95&33.01&10.71&59.66&43.04&63.35&57.02&56.20\\
   CDmamba \cite{cdmamba}   & &11.91&11.91&259.49&12.67&61.20&43.61&59.68&58.88&57.77\\
   CDMask \cite{cdmask}    & &33.52&33.52&128.05& 11.90& 57.42& 36.62&48.46&61.71&49.65\\
   MC-DiSNet \cite{isprs}   & &41.99&41.99&97.98&15.77&61.33&64.86&83.55&73.55&75.47\\
   \hline
   ChangeCLIP~\cite{dong2024changeclip} & \multirow{2}{*}{$\mathcal{M}$} &196.73&133.46&106.00&13.12&60.36&66.65&86.50&73.73&77.80\\
   TextMCD \cite{wu2026textmcd}    & &196.53&48.27&62.01&15.38&61.01&54.35&76.41&69.33&69.30\\
            \hline
   S2M(DINOv3) & \multirow{2}{*}{$\mathcal{S}\rightarrow\mathcal{M}$} & 186.82 & 35.54 & 56.83 & \bf17.80&  \bf 66.14 &  66.39 & \bf88.62 & 71.31 &78.10 \\
   S2M(DINOv3+LoRA) & &187.47&7.71&57.30&16.04&63.90&\bf66.91&87.80&\bf73.85&\bf78.51\\
			\hline
		\end{tabular}
  \vspace{-4mm}
	\end{center}
\end{table*}

\setlength{\tabcolsep}{4pt}
\begin{table*}[ht]
\scriptsize
	\begin{center}
		\caption{
        Performance comparison of RSCD on the LEVIRCD-MM and WHUCD-MM dataset is presented, with all scores reported as percentages and the highest ones highlighted in bold. FLOPs are computed for an input image size of $256 \times 256 \times 3$.
		}
  \vspace{1mm}
		\label{table:whu}
            \begin{tabular}{c||c||ccccc||ccccc}
		\Xhline{1.2pt}
            \rowcolor{mygray}
		    & & \multicolumn{5}{c||}{LEVIRCD-MM} &\multicolumn{5}{c}{WHUCD-MM} \\
            \rowcolor{mygray}
			\multicolumn{1}{c||}{\multirow{-2}{*}{Method}} & \multicolumn{1}{c||}{\multirow{-2}{*}{modality}}
               & F$_1$  & IoU &OA &Pre. &Rec.& F$_1$  & IoU &OA &Pre. &Rec.\\			
                \hline \hline
           
   BIT \cite{BIT} & \multirow{14}{*}{$\mathcal{S}$} &90.23&82.19&99.01&90.34&90.11&90.79&83.13&99.18&93.30&88.41 \\
   ChangeFormer~\cite{changeformer} &&90.40&82.48&99.04&\bf91.67&89.15&90.52&82.68&90.15&92.74&88.40 \\  
   CDxLSTM~\cite{wu2025cdxlstm}      &&90.92&83.35&99.08&91.54&90.30&90.88&83.28&99.14&88.51&\bf93.38 \\  
   DDLNet \cite{ddlnet}   &&90.14&82.05&99.01&91.12&89.18&90.97&83.44&99.19&92.87&89.15 \\  
   SNUNet \cite{snunet}      &&90.27&82.27&99.01&90.59&89.96&77.57&63.36&97.77&71.98&84.10 \\ 
   STNet \cite{STNet}       &&90.64&82.88&99.05&91.40&89.89&88.40&79.22&98.95&89.78&87.07 \\  
   USSFCNet \cite{usscfnet}    &&88.99&80.17&98.86&87.77&90.25&89.97&81.77&99.09&91.35&88.63 \\  
   LGPNet \cite{LGPNet}      &&90.74&83.06&99.06&90.80&90.69&79.41&65.86&97.90&72.21&88.21 \\  
   Changemamba \cite{changemamba} &&90.43&82.54&99.03&91.12&89.76&91.92&85.05&99.29&\bf95.67&88.45 \\  
   RSmamba \cite{RSmamba}     &&89.61&81.18&98.95&90.35&88.89&88.76&79.80&99.01&92.51&85.31 \\    
   CDmamba \cite{cdmamba}     &&89.74&81.39&98.96&90.31&89.18&89.96&81.76&99.12&94.11&86.16 \\
   CDMask \cite{cdmask}      &&90.30&82.32&99.02&91.00&89.61&89.62&81.19&99.04&89.20&90.04 \\
            
   MC-DiSNet \cite{isprs} & &90.72&83.02&99.06&90.94&90.50&92.79&86.55
&99.35	&93.98&91.63
 \\
   \hline
   ChangeCLIP~\cite{dong2024changeclip}  &\multirow{2}{*}{$\mathcal{M}$}&89.90&81.65&98.97&89.69&90.10&87.52&77.80&98.88&89.16&85.93 \\ 
   TextMCD \cite{wu2026textmcd}    & &89.79&80.64&98.89&89.58&90.02&85.63&74.87&98.61&81.70&89.96 \\  
   
   			\hline
   S2M(DINOv3) &\multirow{3}{*}{$\mathcal{S}\rightarrow\mathcal{M}$} & \bf90.98 & \bf83.46 & 99.08 & 90.53&\bf 91.45 &  91.57& 84.46 & 99.25 & 94.19& 89.10 \\
   S2M(DINOv3+LoRA)& & 90.79 & \bf83.46 & 99.06 & 90.52& 91.06 &  \bf 93.49& \bf87.78 & \bf99.41 & 94.84& 92.15 \\
			\hline
		\end{tabular}
  \vspace{-4mm}
	\end{center}
    \label{tabcamp}
\end{table*}

\begin{figure}[htpb]
	\centering
	\includegraphics[width=1\textwidth]{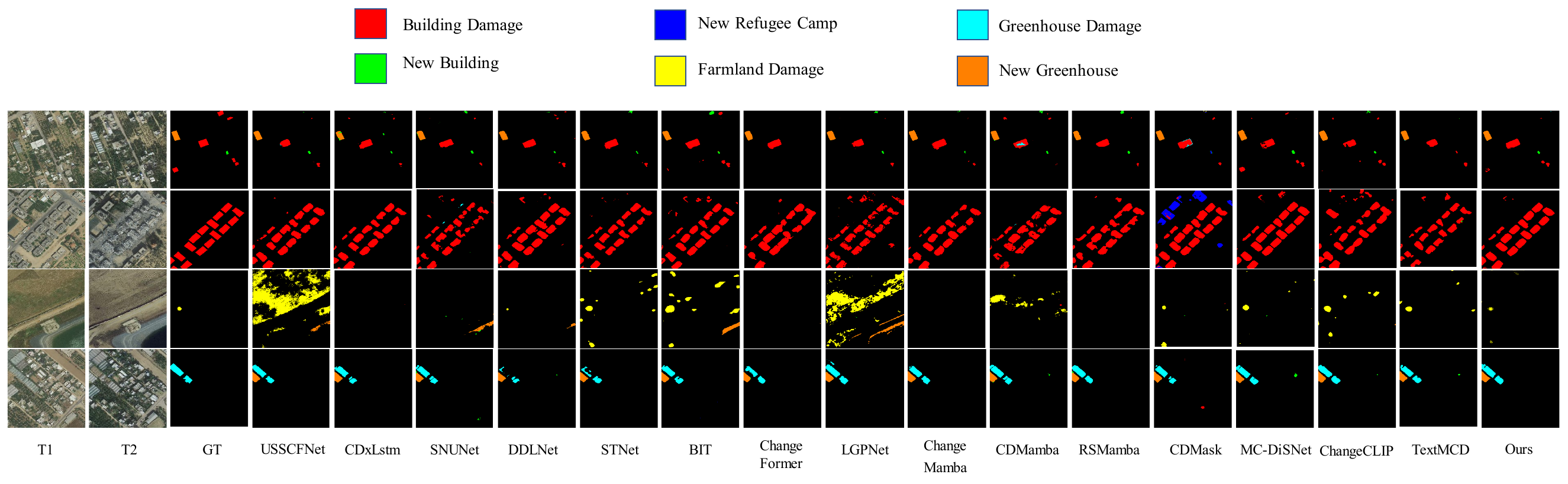}
    \vspace{-5mm}
	\caption{Example results on Gaza-change-v2 dataset.}
	\label{fig:compgaza}
 \vspace{-4mm}
\end{figure}

\begin{figure}[htpb]
	\centering
	\includegraphics[width=1\textwidth]{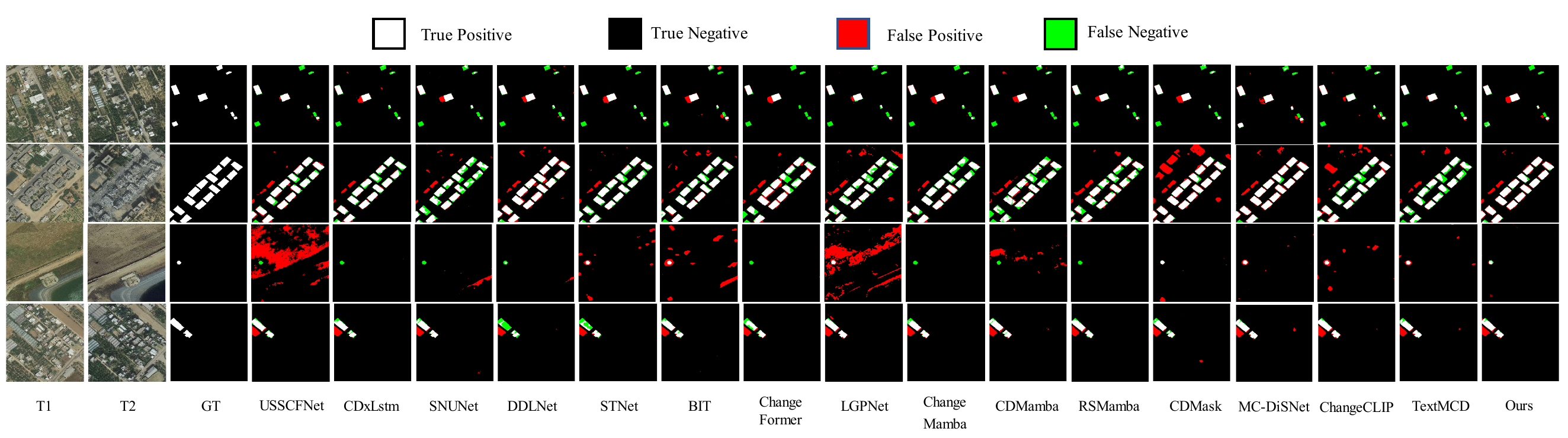}
    \vspace{-8mm}
	\caption{Qualitative comparison on the Gaza-Change-v2 dataset. Color coding: white for true positives, black for true negatives, red for false positives, and green for false negatives.}
	\label{fig:binarycompgaza}
 \vspace{-4mm}
\end{figure}

\textbf{Visual comparison.} Figures~\ref{fig:compgaza} and~\ref{fig:binarycompgaza} present qualitative segmentation results on the Gaza‑Change‑v2‑MM dataset.
Several visual patterns corroborate the quantitative findings and highlight the unique advantages of structured multimodal supervision. The second row depicts a scene containing numerous small‑scale change instances.
Due to the limited pixel extent of each object and the dominance of the unchanged background in the loss function, unimodal methods frequently miss such targets.
In contrast, S2M produces more complete detection results with fewer omissions.
The third row presents a representative case.
When the visual appearances of different change types overlap (e.g., damaged buildings and debris‑covered bare land), unimodal methods tend to produce scattered false detections, as their purely visual features lack the semantic priors necessary to distinguish such confusable patterns.
By comparison, the multimodal method equipped with structured semantic quadruples (S2M) yields significantly fewer false positives.

\textbf{Discussion.} S2M consistently generates more accurate segmentation masks and produces more continuous segmentation boundaries.
These visualizations are consistent with the quantitative results reported above, reinforcing the central claim of this work: structured semantic quadruples distilled from mask labels provide precise and actionable cross‑modal guidance that effectively resolves the inter‑class similarity challenge in remote sensing change detection.

\subsection{Ablation Studies}

As shown in the Table~\ref{tab:text_ablation_v3}, after incorporating text, the three models achieved Recall improvement of 9.62\%, 6.38\%, and 4.45\% respectively, demonstrating that text can effectively reduce omission errors. The mF$_1$ and mIoU of all models increased, indicating that text leads to more precise and complete change detection results. A slight increase in false alarms is compensated for by the substantial gain in recall, leaving the overall performance still superior. This experiment fully proves that in multimodal change detection tasks, fusing textual information can effectively enhance the model's ability to perceive changed regions, with notable advantages particularly in recall and overall robustness. S2M, as a free information lunch, is highly worth applying in unimodal change detection tasks.

\setlength{\tabcolsep}{4pt}
\begin{table*}[htbp]
\centering
\vspace{-4mm}
\scriptsize
\caption{Ablation study on the impact of textual information for change detection on the Gaza-Change-v2-MM dataset. 
         \XSolidBrush = without S2M, \Checkmark = with S2M. 
         \textcolor{blue}{$\uparrow$} indicates improvement after adding S2M. 
         Best per model per metric are in \textbf{bold}.}
\label{tab:text_ablation_v3}
\begin{tabular}{@{} l l c c c c c c @{}}
\toprule
\rowcolor{mygray}
		     Model&Text &Sek &F$_{\text{scd}}$ & mIoU &Pre. &Rec. & mF$_1$ \\	
                \hline \hline
\multirow{2}{*}{CDxLSTM} 
& \XSolidBrush & 15.45 & 62.84 & 60.30 & \bf86.02 & 64.74 & 72.56 \\
& \Checkmark   & \textbf{16.15} \textcolor{blue}{$\uparrow$} & \textbf{63.91} \textcolor{blue}{$\uparrow$} & \textbf{65.56} \textcolor{blue}{$\uparrow$} & 85.72 & \textbf{74.36} \textcolor{blue}{$\uparrow$} & \textbf{77.18} \textcolor{blue}{$\uparrow$} \\
\addlinespace[0.5em]
\multirow{2}{*}{SNUNET} 
& \XSolidBrush & 12.84 & 61.80 & 56.85 &\bf 84.59 & 61.71 & 69.73 \\
& \Checkmark   & \textbf{13.17} \textcolor{blue}{$\uparrow$} & \textbf{62.21} \textcolor{blue}{$\uparrow$} & \textbf{62.67} \textcolor{blue}{$\uparrow$} & 84.33 & \textbf{68.09} \textcolor{blue}{$\uparrow$} & \textbf{74.58} \textcolor{blue}{$\uparrow$} \\
\addlinespace[0.5em]
\multirow{2}{*}{STNET} 
& \XSolidBrush & 16.58 & 64.81 & 64.08 &\bf 87.66 & 70.42 & 76.66 \\
& \Checkmark   & \textbf{16.65} \textcolor{blue}{$\uparrow$} & \textbf{64.96} \textcolor{blue}{$\uparrow$} & \textbf{64.45} \textcolor{blue}{$\uparrow$} & 87.06 & \textbf{74.87} \textcolor{blue}{$\uparrow$} & \textbf{76.89} \textcolor{blue}{$\uparrow$} \\
\midrule
\multicolumn{2}{l}{\textbf{Average improvement ($\Delta$)}} 
& $+0.37$ & $+0.54$ & $+3.83$ & $-0.39$ & $+6.82$ & $+2.96$ \\
\bottomrule
\end{tabular}
\vspace{-2mm}
\end{table*}

In Table \ref{table:att}, we compare the impact of different combinations of four attributes. Using both type and category (row 2) notably improves SEK (17.54\%) and F$_{scd}$ (64.75\%) but reduces mIoU to 68.92\%. When all four attributes are used (row 8), SEK and F$_{scd}$ peak, while mIoU (66.39\%) and mF$_1$ (78.10\%) fall below the no‑attribute baseline. Overall, different attributes exhibit complementarity and trade‑offs. The optimal combination depends on the task objectives.

\setlength{\tabcolsep}{4pt}
\begin{table*}[htpb]
\scriptsize
	\begin{center}
    \vspace{-4mm}
		\caption{
		Ablation experiments with four attributions on the Gaza-Change-v2-MM dataset.
		}
  \vspace{0pt}
		\label{table:att}
            \begin{tabular}{cccc||cccccc}
		\Xhline{1.2pt}
            \rowcolor{mygray}
		      $[$quantity$]$ & $[$type$]$ & $[$category$]$ & $[$location$]$  &Sek &F$_{\text{scd}}$ & mIoU &Pre. &Rec. & mF$_1$ \\	
                \hline \hline
          \XSolidBrush & \XSolidBrush & \XSolidBrush & \XSolidBrush & 14.82 & 60.80 & 71.05 & 86.72 & 79.52 &81.31\\
           \XSolidBrush &\Checkmark & \Checkmark & \XSolidBrush &17.54&64.75&68.92&87.47&78.20&80.08 \\
           \Checkmark &\XSolidBrush & \Checkmark & \XSolidBrush &16.37&62.98&70.00&86.93&78.39&80.55 \\
           \XSolidBrush &\XSolidBrush & \Checkmark & \Checkmark &17.27&64.70&67.04&\bf89.16&69.83&78.06 \\
           \Checkmark &\Checkmark & \XSolidBrush & \XSolidBrush &16.64&63.88&70.48&87.15&\bf79.91&80.83 \\
           \XSolidBrush &\Checkmark & \XSolidBrush & \Checkmark &16.86&63.42&\bf 71.57&88.06&81.17&\bf81.93 \\
           \Checkmark &\XSolidBrush & \XSolidBrush & \Checkmark &16.76&63.11&70.97&87.22&76.51&80.59 \\
   			\hline
            \Checkmark &\Checkmark & \Checkmark & \Checkmark &\bf 17.80&\bf66.14&66.39&88.62&71.31&78.10 \\
            \hline
		\end{tabular}
  \vspace{-4mm}
	\end{center}

\end{table*}


\vspace{-3mm}
\section{Conclusion}
\vspace{-3mm}

This paper proposes S2M, a framework that for the first time automatically extracts structured text in the form of a quadruple (where, what, how, how many) directly from change masks, upgrading unimodal data to multimodal data at zero additional annotation cost. By combining a two-stage training strategy with a bidirectional vision-text contrastive loss, S2M achieves deep cross-modal alignment. On the Gaza-Change-v2, LEVIR-CD, and WHU-CD datasets, S2M consistently attains state-of-the-art performance, effectively mitigating the problem of visually similar but semantically different pseudo-changes, and demonstrating that masks can indeed "talk".

{
\small
\bibliographystyle{plain}
\bibliography{ref.bib}
}

\end{document}